\documentclass[runningheads]{llncs}
\usepackage{graphicx}
\usepackage{amsfonts}
\usepackage{amssymb}
\usepackage{xcolor}
\usepackage{adjustbox}
\usepackage{amsmath}

\begin{document}
\title{Semi-Siamese Network for Robust Change
Detection Across Different Domains with
Applications to 3D Printing}
\titlerunning{Semi-Siamese Change Detection}
%
\author{Yushuo Niu\inst{1} \and
Ethan Chadwick\inst{2} \and
Anson W. K. Ma\inst{2} \and Qian Yang\inst{1}  }
\authorrunning{Y. Niu et al.}
%
\institute{Computer Science \& Engineering Department, University of Connecticut \\
371 Fairfield Way, Unit 4155, Storrs, CT 06269-4155 \\
\email{\{yushuo.niu,qyang\}@uconn.edu} \\
\and Chemical \& Biomolecular Engineering Department, University of Connecticut \\
97 North Eagleville Road, Unit 3136, Storrs, CT 06269-3136\\
\email{\{ethan.chadwick,anson.ma\}@uconn.edu}}
\maketitle              
\begin{abstract}
Automatic defect detection for 3D printing processes, which shares many characteristics with change detection problems, is a vital step for quality control of 3D printed products. However, there are some critical challenges in the current state of practice. First, existing methods for computer vision-based process monitoring typically work well only under specific camera viewpoints and lighting situations, requiring expensive pre-processing, alignment, and camera setups. Second, many defect detection techniques are specific to pre-defined defect patterns and/or print schematics. In this work, we approach the defect detection problem using a novel Semi-Siamese deep learning model that directly compares a reference schematic of the desired print and a camera image of the achieved print. The model then solves an image segmentation problem, precisely identifying the locations of defects of different types with respect to the reference schematic. Our model is designed to enable comparison of heterogeneous images from different domains while being robust against perturbations in the imaging setup such as different camera angles and illumination. Crucially, we show that our simple architecture, which is easy to pre-train for enhanced performance on new datasets, outperforms more complex state-of-the-art approaches based on generative adversarial networks and transformers. Using our model, defect localization predictions can be made in less than half a second per layer using a standard MacBook Pro while achieving an F1-score of more than 0.9, demonstrating the efficacy of using our method for \textit{in-situ} defect detection in 3D printing.

\end{abstract}
\section{Introduction}

\begin{figure*}
    \centering  
     \includegraphics[width=\textwidth]{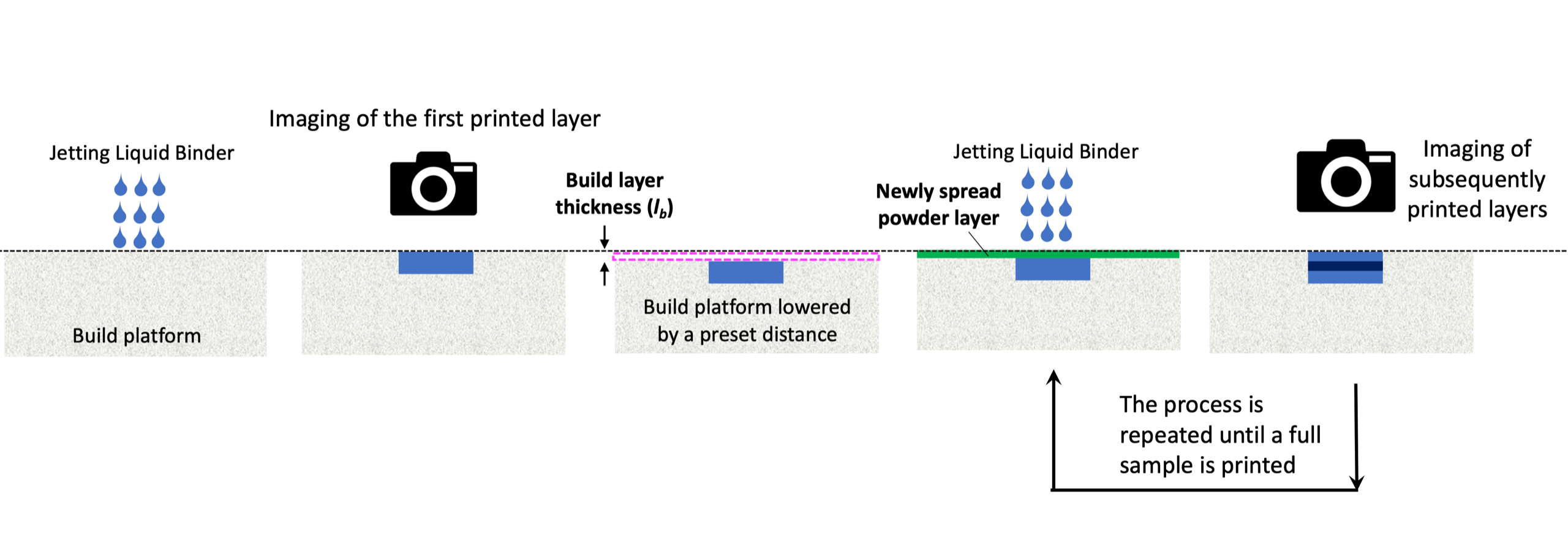}   
    \caption{Schematic diagram of the 3D printing process for binder jet 3D printing with layer-wise imaging during the print.}
    \label{fig:3Dprintschematic}
\end{figure*}

Defect detection methods that can provide feedback in real-time is of significant interest to the additive manufacturing community in order to save on materials cost, printing time, and most importantly, to ensure the quality of printed parts. A key advantage of 3D printing technology that can be leveraged to enable \textit{in situ} defect detection is that 3D objects are printed layer by layer (Figure~\ref{fig:3Dprintschematic}). Thus, each 2D layer of the object can be imaged and probed for internal defects; unlike traditional manufacturing processes, it is not necessary to wait to analyze the fully printed 3D object, and the interior of the object can be probed as the object is being constructed.

\begin{figure}
    \centering
    \includegraphics[width=0.5\textwidth]{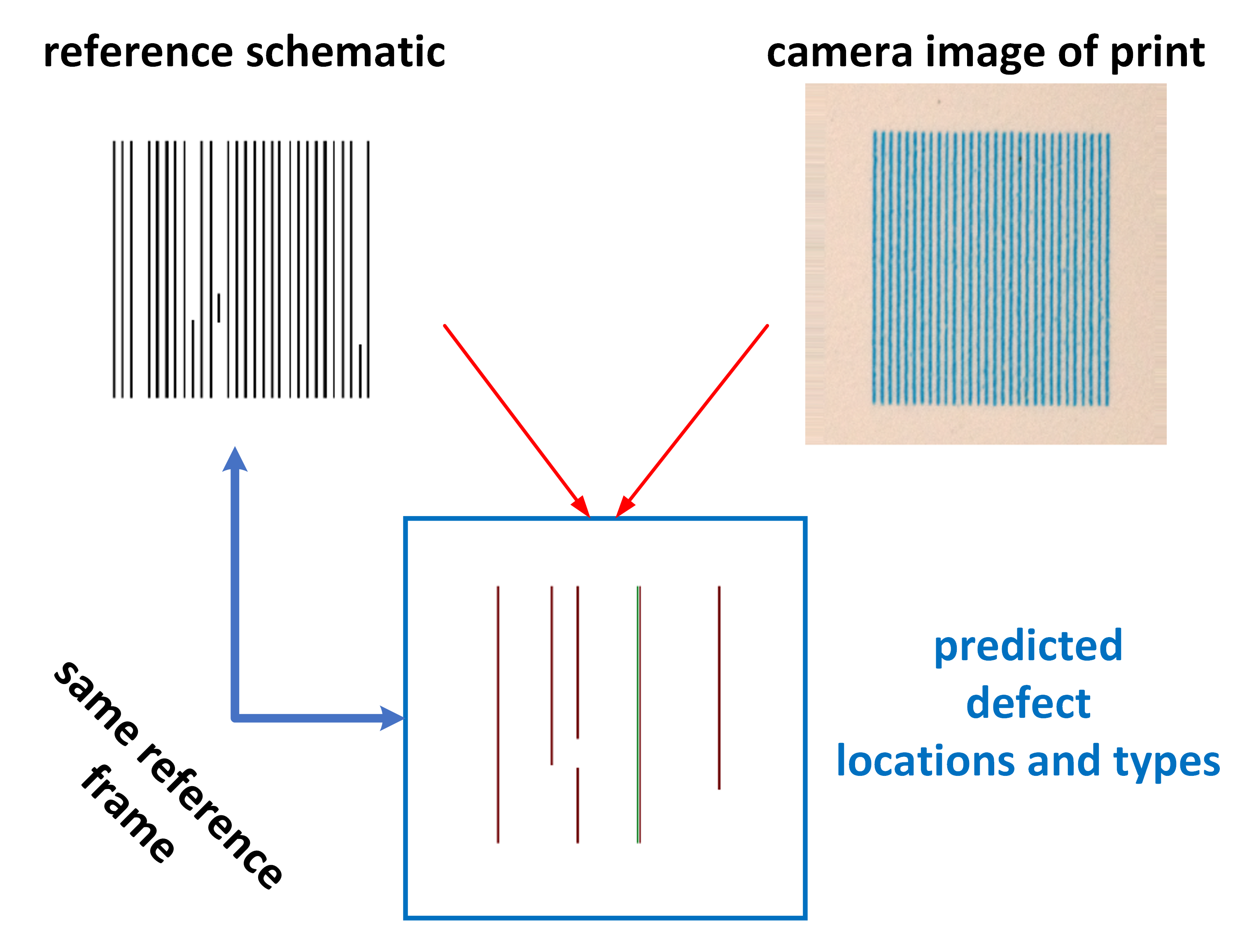}
    \caption{Our robust defect localization model takes as input a reference print schematic and a camera image of the print, and predicts the precise location and type of defects with respect to the frame of reference of the print schematic. In the predicted image, white corresponds to no defects, red to over-extrusion and green to under-extrusion. This model can be used \textit{in-situ} for defect detection: it requires 0.419 seconds for prediction while printing a single layer on an inkjet-based 3D printer requires tens of seconds or less.}
    \label{fig:introschematic}
\end{figure}

In this work, we propose a novel defect detection method for 3D printing that poses the problem as one of change detection between a desired reference schematic and a camera image of the printed layer (Figure~\ref{fig:introschematic}). In the change detection problem in computer vision, two images such as satellite images of land or surveillance images of streets are compared for differences. There are several challenges common to both the defect detection and change detection problems: the need to pre-process and pre-align images due to changes in camera angle and lighting, which result in significant and sometimes expensive limitations to the camera setup that must be used, and the data-hungry nature of this complex comparison problem. Additionally, the images we would like to compare for 3D printing are from heterogeneous sources: one is a reference print schematic and the other is a noisy camera image of the actual printed result. In this work, we utilize one-shot learning techniques~\cite{Koch2015SiameseNN} to develop a novel deep learning architecture that can provide fast and precise localization of defects robust to camera angle and lighting perturbations. A key characteristic of our model is that its relatively lightweight and simple architecture can be easily pre-trained to adapt to new datasets; in fact we show that pre-training enables our simple model to outperform complex state-of-the-art approaches techniques such as transformers. The simplicity and flexibility of our model will enable it to be highly transferable to different industrial settings for 3D printing, without requiring careful camera setups and application-specific model customization that is both expensive and time-consuming. Our proposed approach of building on change detection methods from computer vision for tackling the challenges of defect detection is to the best of our knowledge a new direction in the 3D printing field.

\section{Related work}
Change detection is a fundamental task in computer vision, with many important applications such as analysis of satellite imagery for agricultural monitoring~\cite{Khan2017}, urban planning~\cite{Saha2021}, and disaster assessment~\cite{Sublime_2019}, among others. A large body of work has thus been built starting from at least the 1980s using methods such as change vector analysis~\cite{Malila1980ChangeVA}. To handle perturbations such as misalignment and varied lighting, techniques such as incorporating active camera relocation have been proposed~\cite{Feng2015FineGrainedCD}. Many state-of-the-art methods today are now based on deep learning, ranging from autoencoders to Siamese neural networks to recurrent neural networks, and various combinations thereof~\cite{rs12101688}. Recently, several methods based on combining convolutional neural networks (CNN) with Siamese architectures have been proposed. One of the earlier such methods, ChangeNet, uses a combination of ResNet, fully connected, and deconvolution blocks in its Siamese branches~\cite{Varghese2018ChangeNetAD}. It is designed to handle different lighting and seasonal conditions in the images, but like most existing methods assumes aligned or nearly aligned image pairs. Interestingly, the architecture is different from traditional Siamese architectures in that the deconvolution layers are not required to have the same weights. This is reminiscent of our proposed Semi-Siamese architecture which we will discuss in Section 3; however, we will propose the opposite - the deconvolution layers are the portion of our architecture that are required to share the same weights. Another interesting recent approach uses a Siamese pair of autoencoders~\cite{Mesquita2020FullyCS}, where the change map is generated based on the learned latent representations. However, this method also assumes coregistered images and can only learn approximate change locations in addition to a classification of whether changes have occurred. A recently proposed architecture that enables fast pixel-level change detection is FC-Siam-diff, a fully convolutional encoder-decoder network with skip connections~\cite{Daudt2018FullyCS}. In this model, there are two encoders that share the same architecture and weights, while there is only one decoder. However, this model again assumes coregistered images. Finally, the challenge of dealing with images that are not necessarily coregistered, with differences in lighting, camera viewpoint, and zoom, was recently addressed in Sakurada \& Okatani~\cite{Sakurada2015ChangeDF} and with CosimNet~\cite{Guo2018LearningTM}. The former uses features learned from CNNs trained for large-scale objected recognition in combination with superpixel segmentation to address the problem of change detection in pairs of vehicular, omnidirectional images such as those from Google Street View. The latter, CosimNet, uses the DeeplabV2 model as a backbone and proposes various modifications to the loss function to provide robustness to perturbations~\cite{Guo2018LearningTM}. Nevertheless, both of these methods still assume that the images being compared are from the same domain, e.g. they are both camera or satellite images, rather than from different domains such as a camera image versus a schematic. Recently, heterogeneous change detection has been addressed using generative adversarial networks~\cite{li2021deep} and transformers~\cite{chen2021remote}.

Despite its importance for additive manufacturing, defect detection has traditionally been a challenging task. First, there are many different types of defects that may be of interest, including defects caused by missing jets, inconsistent jets, angled jets, and cracks in powder bed material, just to name a few that are relevant to inkjet-based 3D printing; other technologies such as fused deposition modeling have their own set of defects. Many heuristic-based methods such as computing the entropy of depth maps have consequently been developed to address specific defect types~\cite{Fastowicz2019Objective3P}. In recent years, both classical machine learning methods such as support vector machines utilizing human-engineered features~\cite{Jacobsmhlen2015DetectionOE} and deep learning-based methods utilizing convolutional neural networks have started to be developed to enable more powerful defect detection~\cite{Jin2019AutonomousIC,Scime2020LayerwiseAD}. However, many of these methods require large amounts of labeled experimental data, which is difficult to obtain. They also typically require fixed, high-resolution camera setups, and cannot easily handle differences in camera angle and lighting. For example, one group of methods is based on denoising autoencoders~\cite{Han2020FabricDD}, where the idea is that an autoencoder is trained to take as input a ``noisy" (defective) image and output its non-defective counterpart. Then, differences between the input and output can be used to identify defects. An advantage of this approach is that it does not require a large amount of labeled experimental data; however, unlike change detection approaches which can handle general differences, this approach can only handle a pre-defined range of defects, since it must be trained to be able to remove them from the output.

\vspace{-10pt}
\section{Semi-Siamese defect detection model}
Our proposed model consists of two major components: a novel Semi-Siamese architecture based on U-Net~\cite{RFB15a} branches, and a fully convolutional network (FCN) to reconstruct the final defect detection mask. The input to the model is a pair of 2D images corresponding to a particular layer during 3D printing: the reference schematic images of the desired print pattern, $I_{ref}\in R^{H \times W \times 3}$, and the camera images of the printed result, $I_{cam}\in R^{H \times W \times 3}$. The image pair $(I_{ref}, I_{cam})$ is first fed into a Semi-Siamese network to generate a pair of feature maps $(F_{\textit{ref}}, F_{\textit{cam}})$ of the same dimensions as the input. In contrast to standard Siamese networks and existing Semi-Siamese networks which use different decoders, a simple but key innovation of our architecture is that the feature extraction sections of each branch (encoder) do not share the same weights; only the reconstruction section (decoder) share the same weights. This is important for our defect detection problem, since the camera image and reference schematic come from different domains. Thus in order to compare them, we would first need to use \textit{different} feature extraction functions to transform them both to the same latent feature space, after which we can then reconstruct them both in a comparable reference frame using the \textit{same} reconstruction function. Then, the Euclidean distance is used to calculate the change map. It is important to calculate the change map from the reconstructed images in a comparable reference frame rather than from the latent feature space in order to enable highly precise pixelwise defect localization. The final FCN is then used to fully transform the change map from this comparable reference frame back to the reference frame of the schematic image.

\vspace{-10pt}
\subsection{Transfer learning from U-Net models}
\label{subsec:unet}
    As described above, the Semi-Siamese branches of our model are based on the U-Net architecture. This choice is made to leverage the ability of U-nets to produce high resolution outputs~\cite{RFB15a}, enabling precise localization of defects upon comparison of the outputs $(F_{\textit{ref}}, F_{\textit{cam}})$ from each branch. In order to further improve the performance of our model, we first utilize transfer learning from a U-Net model with the same architecture as our Semi-Siamese branches. This U-net model takes as input a perturbed camera image, and outputs a transformation of the image into the same reference frame as its corresponding reference print schematic. When trained on a fixed number of reference schematics, this U-Net can be used for detecting defects by directly comparing a camera image transformed into the reference frame with its corresponding print schematic. However, it is important to note that this architecture cannot handle arbitrary print schematics. Suppose that we would like to detect defects in a print corresponding to a schematic (called ``schematic-new") that is similar to a print schematic that the model was previously trained on (called ``schematic-old"), but looks like a perturbed version of that schematic. Then any camera images of a perfect print of "schematic-new" might be erroneously transformed by the model back into the reference frame of ``schematic-old". Now when compared with ``schematic-new", many defects will be detected, even though no defects occurred in the actual print. Thus, this U-Net architecture cannot be used on its own to handle defect detection for arbitrary desired print schematics. We will instead use this U-Net model pre-trained on a small set of reference schematics to initialize the weights of each branch of our Semi-Siamese model. This allows us the initialize the Semi-Siamese model in such a way that it can offset perturbations for some limited sets of camera images and reference schematics. We then continue training to fine-tune these weights to be able to handle arbitrary reference print schematics and perturbed camera images. As we show in the results, this ability to pre-train the U-Net to initialize our Semi-Siamese model is key to high performance on novel problems. We note that while we have utilized a U-Net backbone for our Semi-Siamese branches, we can replace it with any state-of-the-art encoder-decoder architecture of choice.

\vspace{-10pt}
\subsection{Semi-Siamese network architecture}
\begin{figure*}
    \centering
    \includegraphics[width=\textwidth]{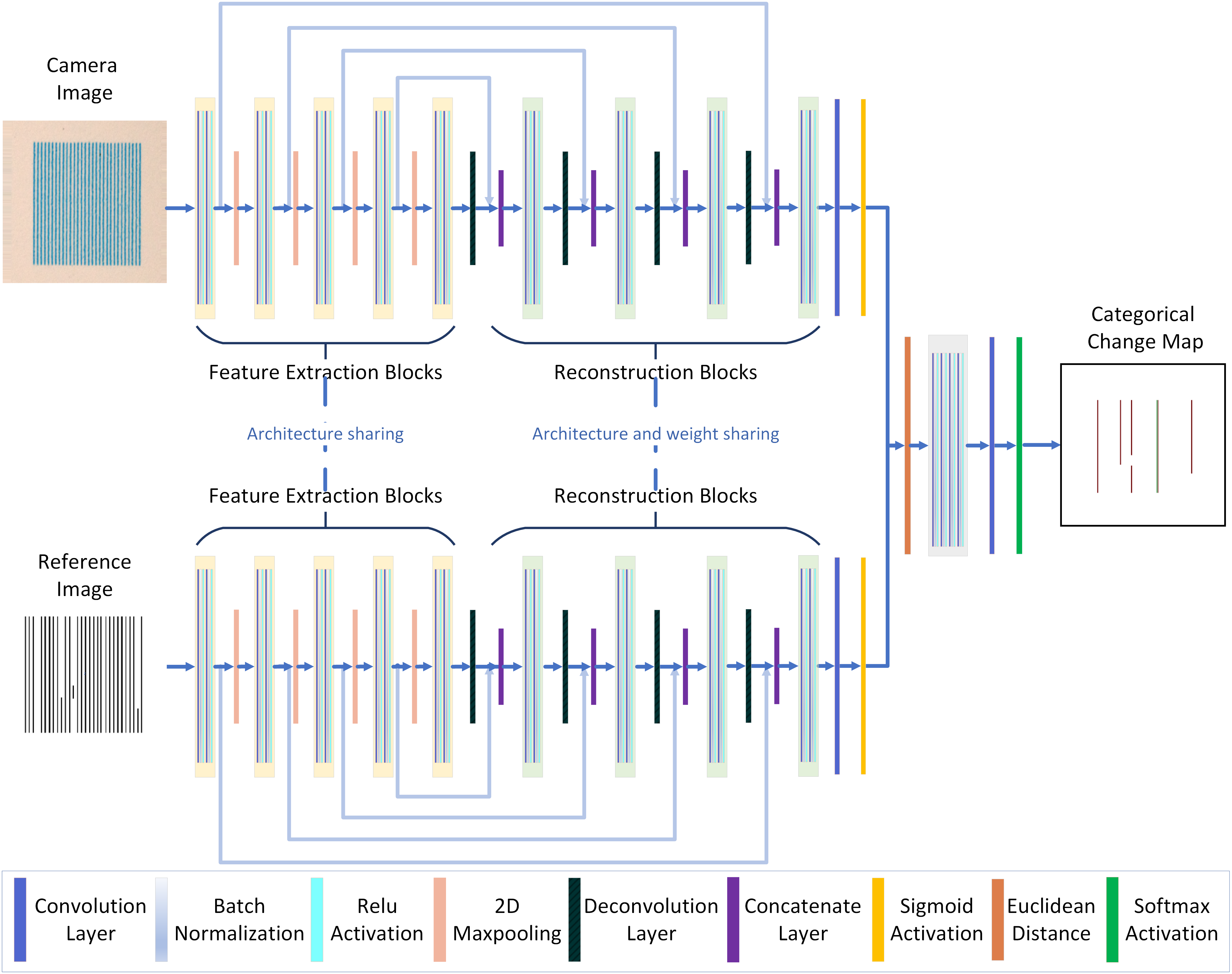}
    \caption{Full architecture for our model. Different types of layers are labeled by color, as indicated in the legend. The Semi-Siamese branches each consist of an underlying U-Net architecture, with a feature extraction (encoder) section, a reconstruction (decoder) section, and skip connections between corresponding layers.}
    \label{fig:semisiam}
\end{figure*}

Our deep learning model begins with two U-Net branches sharing an identical architecture. Each U-Net has five fully convolutional blocks to do downsampling (feature extraction) and four convolutional blocks to do upsampling (reconstruction). Each feature extraction block is composed of two $3 \times 3$ convolutional layers followed by a batchnormalization layer and a rectified linear unit (ReLU) activation. For the first four feature extraction blocks, there is a 2D max pooling layer after each block, where each max pooling layer has pool size $2 \times 2$ and strides of 2. For each of the first four feature extraction blocks, the size of the feature maps is thus reduced by half, while the number of channels is doubled. In the last feature extraction block, there is no max pooling layer, so the size of the feature map remains the same and only the number of channels is doubled. For the reconstruction blocks, each block starts with $3 \times 3$ convolution layers followed by a batchnormalization layer $3 \times 3$ convolution layers with ReLU activation. Analogous to the feature extraction block, the size is doubled each time but the number of channels is reduced by half. Before each reconstruction layer, there is a 2D transposed convolutional layer for upsampling (upsampling layer). Skip connections link the output from the max pooling layers to the corresponding upsampling layers. From these Semi-Siamese branches, a pair of feature maps are generated and their pixel-wise Euclidean distance is calculated to get the change map. This change map is then fed into the remaining FCN, which generates the final change mask $\hat{Y}$ giving the predicted probability of each class (non-defect, over extrusion, and under extrusion) that each pixel corresponds to a defect location. A full schematic of the proposed architecture is shown in Figure~\ref{fig:semisiam}.

\subsection{Training Objective}

Our model uses focal loss~\cite{lin2017focal} as the loss function in order to address the imbalance in change detection datasets between easy-to-classify background pixels and the smaller number of foreground pixels where changes may occur. We note that this imbalance is more pronounced in datasets from our 3D printing application than in benchmark datasets for change detection typically derived from satellite imagery. The focal loss leverages a modulating term to the cross entropy loss in order to focus learning on hard samples by reducing the contributions to the training loss of samples for which the model is confidently correct (easy samples). The equation for the focal loss is:
\begin{equation}
    \mathcal{FL}(p) = -\sum\limits_{i=1}^{N}\alpha_{i} (1-p_{i})^{\gamma} \log (p_{i})
\end{equation}
where $(1-p_{i})^{\gamma}$ is the modulating term, with tunable focusing parameter $\gamma \geq 0$. The hyperparameter $\alpha$ is an additional weighting parameter that re-balances the contribution of each sample to the loss, typically based on the true sample class.

\vspace{-10pt}
\section{Experiments}
    We compare our model against three change detection methods that represent different existing state-of-the-art approaches: ChangeNet~\cite{Varghese2018ChangeNetAD}, which utilizes a ResNet backbone and Semi-Siamese branches with shared encoders; BIT~\cite{chen2021remote}, which utilizes transformers; and DTCDN, which utilizes generative adversarial networks~\cite{li2021deep}.

    First, to compare our model with existing models for heterogeneous change detection across different image domains, we use the benchmark Wuhan dataset~\cite{caltagirone2014torre}, which consists of pairs of optical and synthetic aperture radar (SAR) images. Note that this dataset only involves binary classification and does not contain significant perturbations in image alignment and angle; there is no benchmark dataset for heterogeneous change detection in the multi-class case. To test the application of our model to 3D printing with three-class classification (no defect, under-extrusion, and over-extrusion), and also to demonstrate our model's robustness to perturbations in camera angle and lighting, we then created our own experimental dataset consisting of pairs of (1) reference print schematics and (2) top-down camera images of inkjet-based 3D printing on powder bed material, where Dimatix Blue model fluid is used on Visijet core powder. To simplify dataset generation, our set of reference schematic images consist only of images with vertical lines of varying length spaced closely together. The corresponding camera images are taken at various angles, so that a simple template matching approach would not be able to easily achieve pixel-wise accuracy in defect localization. In future work, we will expand to more complex schematic images. However, the efficacy of our model in precisely localizing defects can be sufficiently demonstrated using this dataset.

\subsection{Full dataset generation using data augmentation}
Since generating an experimental dataset is time-consuming, and with a real 3D printer also incurs significant material costs, we start with only a limited dataset of 57 pairs of experimental images. We then use data augmentation to significantly increase our dataset size by adding perturbations in camera angle to existing camera images. Note that lighting perturbations come naturally from the camera images being taken with no special lighting setup. We do not make any changes to the reference schematics. The types of perturbations we use in the data augmentation includes zoom, rotation, shear, and position (width and height) shift. We emphasize that here we use data augmentation to create our initial full dataset, in contrast to the typical setting in computer vision where data augmentation should only be used in training. The perturbations to the camera image given by data augmentation correspond to artificial new ``experiments" of different camera setups. To prevent data leakage, we separate the training, validation, and test sets by reference schematic. The final dataset consists of 16400 training, 560 validation, and 560 test images, where 41 underlying schematic images are used in the training data, and 8 schematic images each were used for validation and test.

To create defective image pairs, we match the camera image from one reference image with a different reference schematic. Since all of the images correspond to perfect non-defective prints for their true corresponding schematic, we can precisely localize the ``defects" in the defective image pairs by comparing the camera image's true reference schematic with the given new schematic. To generate the training, validation, and test sets, we first randomly select either one underlying schematic for non-defective examples, or two different underlying schematics for defective examples. Then we randomly pick corresponding camera images from among the perturbed variations in our augmented dataset. We note that it is important to balance the dataset between defective and non-defective pairs; otherwise the trained model tends to predict the presence of some defects even for non-defective pairs. 


\vspace{-10pt}
\section{Results}

Table~\ref{tab: Generated vertical line results} shows the results of different change detection methods, as well as an ablation study on the Semi-Siamese and transfer learning components of our approach, on our generated vertical line dataset. Note that due to the imbalance between classes in this dataset, the under-extrusion class was the most difficult to correctly identify. We report the F1-score for each class, as well as the averaged macro F1-score. Our Semi-Siamese model with initialization was able to achieve significantly higher performance than the other methods on identifying under-extrusion. Even without initialization, using all of the same hyperparameters for handling the imbalanced dataset, the Semi-Siamese model outperformed other methods included BIT. Figure~\ref{tab: Generated vertical line results}, we provide both visual and quantitative comparisons of each method on several example pairs of images. Compared to existing methods, in most cases our model is able to capture the defect locations more precisely and with less noise.

Table~\ref{tab:whu results} shows the results of different change detection methods (including the ablation study similarly to above) on the benchmark Wuhan dataset, and Figure~\ref{fig:wuhan_prediction} provides visual and quantitative comparisons of each method on sample pairs of images. We note that visually, our Semi-Siamese model with initialization is able to better reproduce the smooth shape of the ground truth mask. In the second case where the macro F1-score of DTCDN and BIT outperform our method, note the key difference is the detection of a change in a round region in the top left quadrant. While this is not labeled in the ground truth, on close inspection one can see that this is not necessarily inconsistent with the SAR image.

In addition to being more accurate, our method is comparatively simpler and more lightweight. From Table~\ref{tab:time}, we can see that our model takes signficantly less time to train than ChangeNet, DTCDN, and BIT.

\vspace{-10pt}
\section{Conclusions}
We have developed a new deep learning-based method for change detection where (i) a camera image is being compared against a desired schematic rather than another camera image, and (ii) perturbations to the camera angle and lighting do not need to be pre-corrected, and coregistration is not necessary. This novel Semi-Siamese model can be applied to obtain precise \textit{in-situ} pixel-wise defect localization for each layer of a 3D print, enabling rapid detection of internal defects, ensuring the quality of 3D printed parts and saving time and material costs. While an acknowledged limitation of this method is that it does not directly handle defects in the z-direction in a \textit{single} layer, due to the ability to observe each printed layer at various perturbed camera angles, large z-direction defects in the top layer will likely project onto the 2D camera image in such a way as to appear as in-plane defects. Robust handling of these types of defects will be explored in future work.

Defect detection for 3D printing is an important industrial challenge that to the best of our knowledge is being addressed with change detection techniques for the first time in this work. The key benefit of utilizing the change detection framework is that it is not necessary to pre-define the desired print schematic, nor to have a large set of annotated data for each defect type. Our model is capable of detecting defects in a few seconds with more than 90\% accuracy, and performs better than many different more complicated state-of-the-art approaches: ResNet-based Semi-Siamese models with shared encoders (ChangeNet), generative adversarial network (GAN)-based models (DTCDN), and transformer-based models (BIT), on not only our custom 3D printing dataset but also on the benchmark heterogeneous change detection Wuhan dataset. The simplicity of our model makes it possible to easily achieve good performance on new problems - it is only necessary to pre-train a U-Net (or other encoder-decoder backbone) and then transfer learn from that onto the Semi-Siamese architecture. The robustness of our algorithm to camera angle and lighting perturbations while enabling domain adaptation, as well as its lower training data requirements, will enable flexibility for utilizing this model in different industrial settings.

\begin{table}
\caption{Performance comparison on the vertical line dataset. Note that due to the imbalance between classes in this dataset, the under-extrusion class was the most difficult to correctly identify.}
\begin{adjustbox}{width=\columnwidth,center}
\begin{tabular}{||c| c| c| c| c| c||} 
 \hline
  \textbf{Method} &  \textbf{Accuracy} & \begin{tabular}{@{}c@{}}\textbf{macro} \\ \textbf{F1-score} \end{tabular} & \begin{tabular}{@{}c@{}}\textbf{no-defect} \\ \textbf{F1-score} \end{tabular} &  \begin{tabular}{@{}c@{}}\textbf{over-extrusion} \\ \textbf{F1-score} \end{tabular} &  \begin{tabular}{@{}c@{}}\textbf{under-extrusion} \\ \textbf{F1-score} \end{tabular} \\
 \hline\hline
 ChangeNet           &0.9842  &0.5646  &0.9920  &0.6431  &0.0588  \\ 
 \hline
 DTCDN               &0.9572  &0.7613  &0.9772  &0.6970  &0.6097 \\
 \hline
 BIT                 & 0.9957  & 0.9267  & 0.9978  & 0.9555  & 0.8268 \\
 \hline
 Unet                &0.9774  &0.7155  & 0.9884 &0.6859  &0.4722  \\
 \hline
 Semi-Siam (w/o init)  &0.9962  &0.9406  &0.9981  &\textbf{0.9709}   &0.8529 \\  
 \hline
 Siamese (w/ init)          &0.9658  &0.6652  &0.9823  &0.4629   &0.5503  \\ 
 \hline
 Semi-Siam (w/ init)     & \textbf{0.9972}  & \textbf{0.9517}  & \textbf{0.9986}  &0.9503   &\textbf{0.9061} \\  
 \hline
\end{tabular}
\end{adjustbox}
\label{tab: Generated vertical line results}
\end{table}

\begin{figure}
    \centering
    \includegraphics[width=0.9\textwidth]{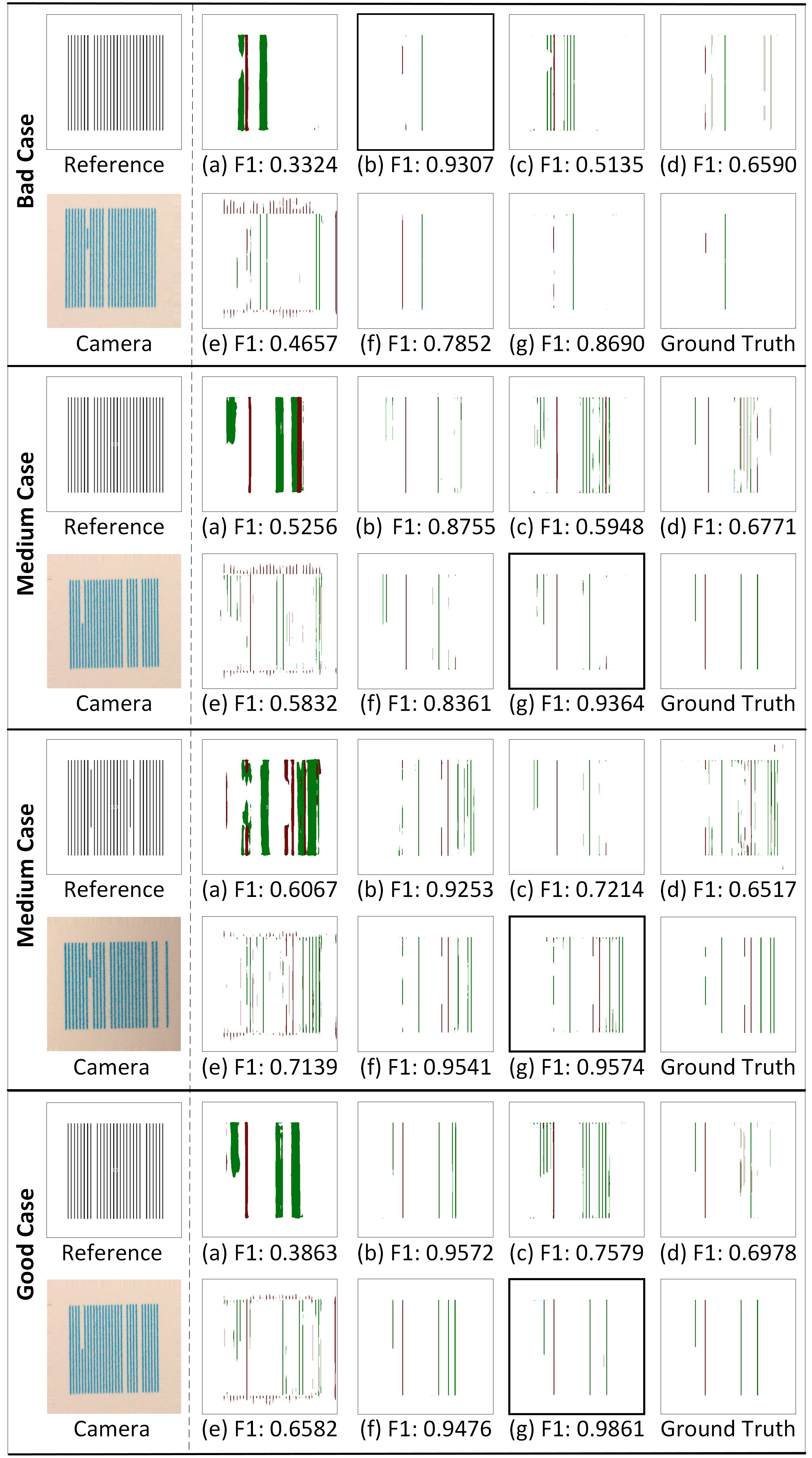}
    \caption{Visual and quantitative (macro F1-score) comparison of different models on the vertical line dataset. The methods from (a)-(g) are (a) ChangeNet, (b) BIT, (c) DTCDN, (d) U-Net, (e) Siamese model with initialization, (f) Semi-Siamese model without initialization, (g) Semi-Siamese model with initialization. Here the ``Ground Truth" column corresponds to the true pixel-wise locations of defects in comparison to the desired ``Reference" image. White corresponds to no defects, red corresponds to over-extrusion, and green corresponds to under-extrusion.}
    \label{fig:wuhan_prediction}
\end{figure}

\begin{table}
\caption{Performance comparison on the Wuhan dataset~\cite{caltagirone2014torre}. DTCDN results are from Ref.~\cite{zhang2022domain} since it was difficult to reproduce the high performance without modifications, as also noted in~\cite{zhang2022domain}}
\begin{center}
\begin{tabular}{||c| c| c| c| c||} 
 \hline
 \textbf{ Method} & \textbf{ Precision} &  \textbf{Recall} &  \textbf{IOU} &  \textbf{F1-score}  \\
 \hline\hline
 ChangeNet            & 0.6555  & 0.6326  & 0.5232 & 0.6420 \\ 
 \hline
 DTCDN$^{\ast}$             & 0.6742 & 0.6536 & 0.5492 & 0.6629 \\
 \hline
 BIT                 & 0.6678 & 0.6859 & 0.5564  &0.6759  \\
 \hline
 Semi-Siam (w/ init)  & 0.6714 & 0.7247 &0.5659  & 0.6905 \\  
 \hline
 Siamese (w/ init)          & 0.6571 &0.6833  & 0.5476  &0.6681  \\ 
 \hline
 Semi-Siam (w/ init)     & \textbf{0.7306} & \textbf{0.7263}  & \textbf{0.6113}  & \textbf{0.7284} \\  
 \hline
\end{tabular}
\end{center}
\label{tab:whu results}
\end{table}

\begin{figure}
    \centering
    \includegraphics[width=0.7\textwidth]{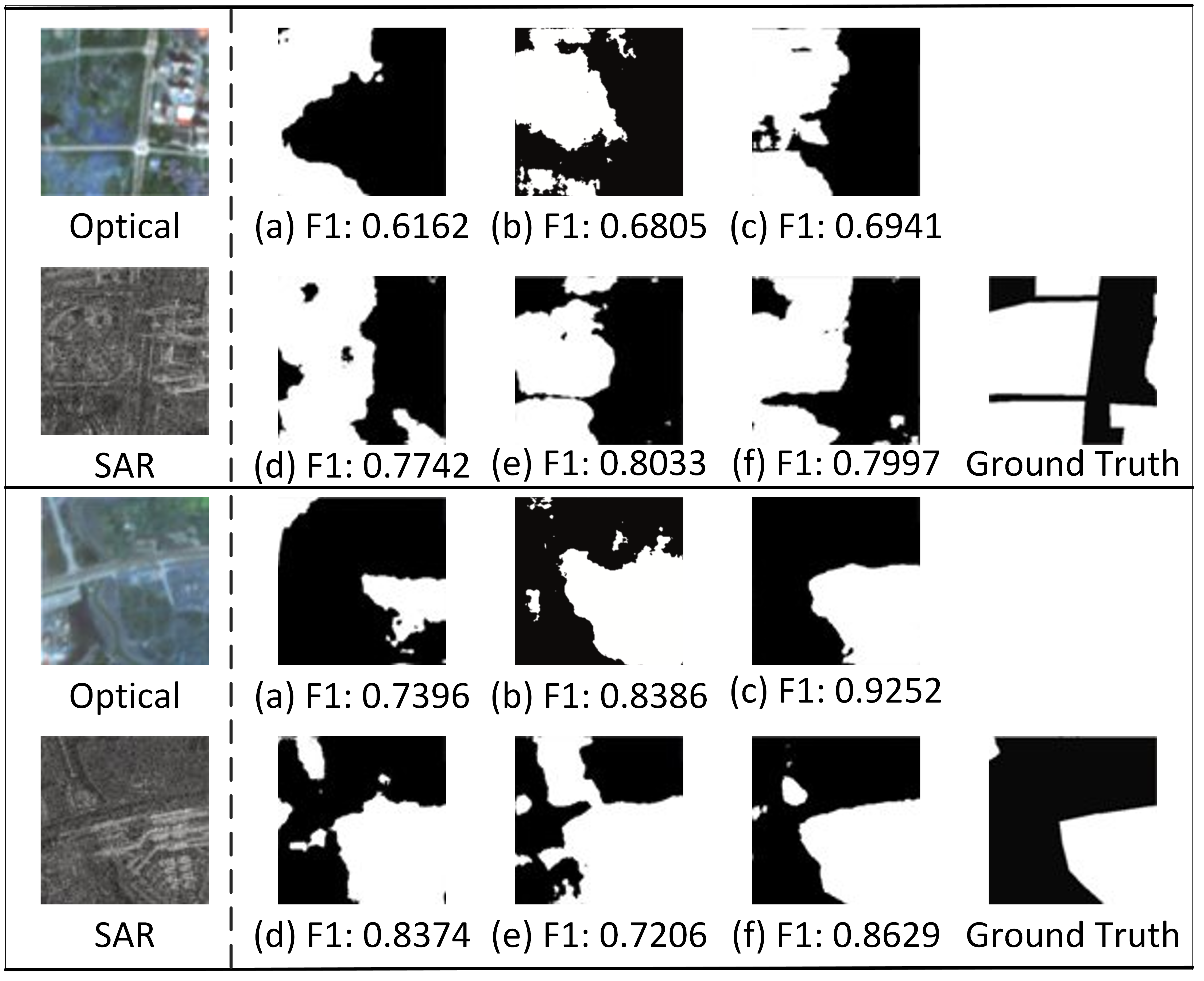}
    \caption{Visual and quantitative comparison of different models on Wuhan dataset~\cite{caltagirone2014torre}. The methods from (a)-(f) are (a) ChangeNet, (b) DTCDN, (c) BIT, (d) Siamese model with initialization, (e) Semi-Siamese model without initialization, Semi-Siamese model with initialization. DTCDN results are from Ref.~\cite{zhang2022domain} as it was difficult to reproduce the high performance without modifications, as also noted in~\cite{zhang2022domain}.}
    \label{fig:wuhan_prediction}
\end{figure}

\begin{table}
\caption{All times are based on training 200 epochs (not including pre-training the GAN in the case of DTCDN, and pre-training the U-Net in the case of Semi-Siamese with initialization) on the vertical line dataset on a 4-GPU workstation.}
\begin{adjustbox}{width=\columnwidth,center}
\begin{tabular}{||c| c| c| c||} 
 \hline
 \textbf{ Method} &  \textbf{Training} &  \textbf{Setup} & \textbf{Prediction} \\
 \hline\hline
 ChangeNet           & 178.14 h  &1797.155 ms  & 423.406 ms\\ 
 \hline
 DTCDN               & \begin{tabular}{@{}c@{}}GAN:36.74 h \\ U-Net++:56.28 h \end{tabular}    & \begin{tabular}{@{}c@{}}GAN: 
1577.17 ms \\ U-Net++:150.958 ms \end{tabular}   & \begin{tabular}{@{}c@{}}GAN:10410.228 ms \\ U-Net++:490.128 ms \end{tabular}  \\
 \hline
 BIT                 & 105.39 h & 569.835 ms  & 342.935 ms  \\
 \hline
 U-Net                &19.34 h & 136.563 ms & 213.76 ms \\
 \hline
 Semi-Siam (w/o init)  & 48.31 h &206.847 ms &419.283 ms  \\  
 \hline
 Siamese (w/ init)          & 36.83 h & 179.195 ms  &408.974 ms \\ 
 \hline
 Semi-Siam (w/ init)     & \begin{tabular}{@{}c@{}}U-Net:9.34 h \\ Semi-Siam: 48.31 h \end{tabular}   &206.847 ms  &419.283 ms  \\  
 \hline
\end{tabular}
\end{adjustbox}
\label{tab:time}
\end{table}
%
%
%
\bibliographystyle{splncs04}
\bibliography{bibnotes}
\end{document}